\documentclass[runningheads]{llncs}
\usepackage{graphicx}
\usepackage[american]{babel}

\usepackage{mathtools} 
\usepackage{algorithm}
\usepackage{algorithmic}
\usepackage{amsmath,amsfonts,amsmath,amssymb}
\usepackage{multirow}
\usepackage{array}
\usepackage{textcomp}
\usepackage{stfloats}
\usepackage{booktabs}
\usepackage{float}
\begin{document}
\newtheorem{myDef}{Definition}
\title{Adversarial Ensemble Training by Jointly Learning Label Dependencies and Member Models
}
\titlerunning{Adversarial Ensemble Training}

\author{Lele Wang\inst{1}\thanks{work done when he was with Research Center for Applied Mathematics and Machine Intelli-gence, Zhejiang Lab.}\and
Bin Liu\inst{2}\thanks{Correspondence author. This paper was accepted by 19th Inter. Conf. on Intelligent Computing (ICIC 2023).}
}
\authorrunning{L. Wang et al.}
%
\institute{Research Center for Data Mining and Knowledge Discovery, Zhejiang Lab,
Hangzhou, 311121 China
\and Research Center for Applied Mathematics and Machine Intelligence,\\
Zhejiang Lab, Hangzhou, 311121 China \\
\email{\{wangll,liubin\}@zhejianglab.com}}
\maketitle              
\begin{abstract}
Training an ensemble of diverse sub-models has been empirically demonstrated as an effective strategy for improving the adversarial robustness of deep neural networks. However, current ensemble training methods for image recognition typically encode image labels using one-hot vectors, which overlook dependency relationships between the labels. In this paper, we propose a novel adversarial en-semble training approach that jointly learns the label dependencies and member models. Our approach adaptively exploits the learned label dependencies to pro-mote diversity among the member models. We evaluate our approach on widely used datasets including MNIST, FashionMNIST, and CIFAR-10, and show that it achieves superior robustness against black-box attacks compared to state-of-the-art methods. Our code is available at \url{https://github.com/ZJLAB-AMMI/LSD}.
\end{abstract}
\section{Introduction}
\label{sec:INTRODUCTION}
Deep neural networks (DNNs), also known as deep learning, have achieved remarka-ble success across many tasks in computer vision \cite{he2016deep,krizhevsky2012imagenet,russakovsky2015imagenet,szegedy2016rethinking}, speech recognition \cite{graves2014towards,hannun2014deep}, and natural language processing \cite{sutskever2014sequence,young2018recent}. However, numerous works have shown that modern DNNs are vulnerable to adver-sarial attacks \cite{goodfellow2014explaining,papernot2016limitations,carlini2017towards,madry2017towards,xiao2018generating,xiao2018spatially}. Even slight perturbations to input images, which are imperceptible to humans, can fool a high-performing DNN into making incorrect predictions. Additionally, adver-sarial attacks designed for one model may deceive other models, resulting in wrong predictions - this issue is known as adversarial transferability \cite{papernot2016transferability,liu2016delving,inkawhich2020transferable,ilyas2019adversarial}. These adversarial vulnerability issues pose significant challenges for real-life applica-tions of DNNs, particularly for safety-critical problems such as self-driving cars \cite{maqueda2018event,bojarski2016end}. As a result, there has been increasing attention on promoting robustness against ad-versarial attacks for DNNs.

\begin{figure}[!h]
\centering
\includegraphics[width=3.5in]{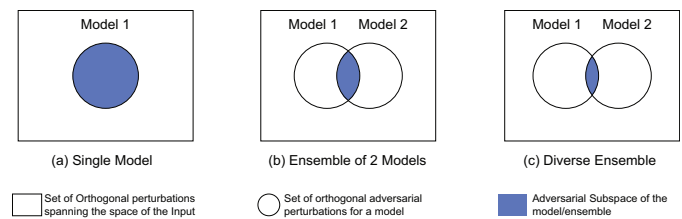}
\caption{A conceptual illustration of the idea of using model ensembling to promote adversarial robustness: (a) single model; (b) an ensemble of two models; (c) an ensemble of two more diversified models. The shaded region denotes a subspace, adversarial attacks within which can fool the (ensemble) model to make a wrong prediction. This figure is drawn referring to Fig.1 of \cite{kariyappa2019improving}.}
\label{fig:adversarial_subspace}
\end{figure}

Ensembling models has been shown to be a highly effective approach for improving the adversarial robustness of deep learning systems. The basic idea behind model ensembling is illustrated through a Venn diagram in Fig.\ref{fig:adversarial_subspace} \cite{kariyappa2019improving}. The rectangle represents the space spanned by all possible orthogonal perturbations to an input instance, while the circle represents the subspace spanned by perturbations that are adversarial to the associated model. The shaded region represents the perturbation subspace that is adversarial to the model ensemble. In the case of a single model as shown in Fig.\ref{fig:adversarial_subspace}(a), any perturbation within the circle results in misclassification of the input image. However, for cases that employ an ensemble of two or more models (Fig.\ref{fig:adversarial_subspace}(b) and (c)), successful adversarial attacks require perturbations only within the shaded region, meaning that the attack must fool all individual models in the ensemble. Therefore, promoting diversity among individual models is an intuitive strategy to improve the adversarial robustness of a model ensemble as the less overlap there is among their corresponding adversarial subspaces, the greater the diversity of the individual models. The amount of overlap determines the dimensionality of the adversarial sub-space \cite{kariyappa2019improving}. Throughout this paper, we use the terms individual model, sub-model, and member model interchangeably.

One challenge in applying model ensembling is how to promote diversity among member models while maintaining their prediction accuracy, particularly for non-adversarial inputs. The question becomes: how can we balance the trade-off between diversity and prediction quality during ensemble training? To address this issue, several advanced methods have been proposed in the literature \cite{pang2019improving,kariyappa2019improving,sen2020empir,zhang2020diversified,yang2021trs}. For example, Pang et al. propose a diversity regularized cost function that encourages diversity in the non-maximal class predictions given by the last soft-max layers of the member models \cite{pang2019improving}. Kariyappa \& Qureshi select an ensemble of gradient misaligned models by minimizing their pairwise gradi-ent similarities \cite{kariyappa2019improving}. Yang et al. find that merely encouraging misalignment between pairwise gradients is insufficient to reduce adversarial transferability \cite{yang2021trs}, and thus propose promoting both gradient misalignment and model smoothness. Sen et al. propose training an ensemble of models with different numerical precisions, where models with higher numerical precisions aim for prediction accuracy, while those with lower numerical precisions target adversarial robustness \cite{sen2020empir}.

As previously mentioned, to apply state-of-the-art (SOTA) ensemble training meth-ods, one must select a diversity metric to measure the diversity of member models. This metric may be the difference in non-maximal predictions given by the last soft-max layers of the member models \cite{pang2019improving}, or the difference in gradient vectors associated with the member models \cite{kariyappa2019improving}. It is worth noting that all these training methods use one-hot vectors to encode image classes, meaning each image in the training set is assigned a hard label. However, such hard labels cannot reflect any possible dependency relationships among image classes. Given an image, it is likely that dependency relationships exist between its associated classes. For example, in a handwritten digit image dataset, the number ``0" may look more similar to ``9" than to ``4"; ``3" may look more similar to ``8" than to ``7", and so on. Conditional on a noisy input image, e.g., one whose ground truth label is ``0", there should exist a dependency relationship between labels ``0" and ``9". Using hard labels omits such conditional dependency relationships.

Motivated by the above observation, we propose a novel ensemble training approach called Conditional Label Dependency Learning (CLDL) assisted ensemble training. Our approach jointly learns the conditional dependencies among image classes and member models. Compared to existing methods, our approach selects a different diversity metric that considers the difference in pairwise gradient vectors and predicted soft labels given by member models. The learned soft labels encode dependency relationships among the original hard labels. We find that our approach is more robust against black-box attacks compared to several state-of-the-art methods. In summary, the main contributions of this work are:
\begin{itemize}
  \item We propose a novel diversity metric for adversarial ensemble training that incorporates information from both the gradient vectors associated with member models and predicted soft labels given by member models.
  \item We adapt a label confusion learning (LCL) model developed in \cite{guo2020label} to generate soft labels of images in the context of adversarial ensemble training, originally used for enhancing text classification..
  \item We propose a CLDL-assisted ensemble training algorithm and demonstrate that it complements existing ensemble training methods. In particular, we show that our algorithm is more robust against black-box attacks compared to several state-of-the-art methods.
\end{itemize}

The remainder of this paper is organized as follows. In Section \ref{sec:BACKGROUND}, we present some preliminary information. In Section \ref{subsec:Our approach}, we describe our CLDL-based ensemble training method in detail. In Section \ref{sec:Simulation}, we present the experimental results. We conclude the paper in Section \ref{sec:CONCLUSION}.
\section{Preliminary}
\label{sec:BACKGROUND}
In this section, we present the preliminary knowledge that is related to our work.
\subsection{Notations}
Here we consider a DNN based image recognition task, which involves $C$ classes. Following \cite{yang2021trs}, let $\mathcal{X}$ denote the input space of the DNN model, $\mathcal{Y} = \{1, 2, . . . , C\}$ the class space. A DNN model is trained to yield a mapping function $\mathcal{F}: \mathcal{X} \to \mathcal{Y}$. Let $x$ denote a clean image and $x^{\prime}$ an adversarial counterpart of $x$. Let $\epsilon$ be a pre-set attack radius that defines the maximal magnitude of an adversarial perturbation. That says, for any adversarial perturbation, its $L_p$ norm is required to be less than $\epsilon$. Let $\ell_{\mathcal{F}}(x, y)$ denote the cost function used for training the model ${\mathcal{F}(x, \theta)}$, where $\theta$ denotes the parameter of the model.
\subsection{Definitions}
\begin{myDef}
\textbf{Adversarial attack}\cite{yang2021trs}. Given an input $x \in \mathcal{X}$ with true label $y \in \mathcal{Y}$, $ F(x) = y $. (1) An untargeted attack crafts $\mathcal{A}_{U}(x)=x+\delta$ to maximize $\ell_{\mathcal{F}}(x+\delta,y)$ where $\left \| \delta \right \|_{p}\le \epsilon$. (2) A targeted attack with target label $y_t \in \mathcal{Y}$ crafts $\mathcal{A}_{T}(x)=x+\delta$ to minimize $\ell_{\mathcal{F}}(x+\delta, y_t)$ where $\left \| \delta \right \|_{p}\le \epsilon$ and $\epsilon$ is a pre-defined attack radius that limits the power of the attacker.
\end{myDef}

\begin{myDef}
\textbf{Alignment of loss gradients}\cite{yang2021trs,kariyappa2019improving}. The alignment of loss gradients between two differentiable loss functions $\ell_{\mathcal{F}}$ and $\ell_{\mathcal{G}}$ is defined as:

\begin{equation}
\small
CS(\nabla_{x} \ell_{\mathcal{F}},\nabla_{x} \ell_{\mathcal{G}}) = \frac{\nabla_{x} \ell_{\mathcal{F}}(x, y) \cdot \nabla_{x} \ell_{\mathcal{G}}(x, y)}{\left\|\nabla_{x} \ell_{\mathcal{F}}(x, y)\right\|_{2} \cdot\left\|\nabla_{x} \ell_{\mathcal{F}}(x, y)\right\|_{2}}\label{eq:Alignment of loss gradients}
\end{equation}
which is the cosine similarity between the gradients of the two loss functions for an input $x$ drawn from $ \mathcal{X}$ with any label $y \in \mathcal{Y}$. If the cosine similarity of two gradients is -1, we say that they are completely misaligned.
\end{myDef}
%
%
%
%
\subsection{Adversarial Attacks}
Adversarial attacks aim to create human-imperceptible adversarial inputs that can fool a high-performing DNN into making incorrect predictions. These attacks are typically divided into two basic classes: white-box attacks, which assume the adversary has full knowledge of the model's structures and parameters, and black-box attacks, which assume the adversary has no access or knowledge of any information regarding the model. Here we briefly introduce four typical white-box attacks in-volved in our experiments while referring readers to review papers \cite{ren2020adversarial,yuan2019adversarial,xu2020adversarial} and references therein for more information on adversarial attacks.

\textbf{Fast Gradient Sign Method (FGSM)} FGSM is a typical white-box attacking method to find adversarial examples. It performs a one-step update along the direction of the gradient of the adversarial loss. Specifically, it generates an adversarial example $x^{\prime}$ by solving a maximization problem as follows \cite{goodfellow2014explaining}:
\begin{equation}
x^{\prime}=x+\varepsilon \cdot \operatorname{sign}\left(\nabla_{x} \ell(x, y)\right),
\label{eq:FGSM}
\end{equation}
where $\varepsilon$ denotes the magnitude of the perturbation, $x$ the original benign image sample, $y$ the target label of $x$, and $\nabla_{x} \ell(x, y)$ the gradient of the loss $\ell(x, y)$ with respect to $x$.

\textbf{Basic Iterative Method (BIM)} BIM is an extension of FGSM, which performs FGSM iteratively to generate an adversarial example as follows \cite{kurakin2016adversarial}:
\begin{equation}
\small
x_{i}^{\prime}=\operatorname{clip}_{x, \epsilon}\left(x_{i-1}^{\prime}+\frac{\epsilon}{r} \cdot \operatorname{sign}\left(g_{i-1}\right)\right)
\label{eq:BIM}
\end{equation}
where $x_{0}^{\prime}\triangleq x$, $r$ is the number of iterations, $\operatorname{clip}_{x, \epsilon}\left(A \right)$ is a clipping function that projects $A$ in a $\epsilon$-neighbourhood of $x$, and $g_{i}\triangleq\nabla_{x} \ell\left(x_{i}^{\prime}, y\right)$.

\textbf{Projected Gradient Descent (PGD)} PGD \cite{madry2017towards} is almost the same as BIM, the only difference being that PGD initialize $x_{0}^{\prime}$ as a random point in the $\epsilon$-neighbourhood of $x$.

\textbf{Momentum Iterative Method (MIM)} MIM is an extension of BIM. It updates the gradient $g_{i}$ with the momentum $\mu$ as follows \cite{dong2018boosting}:

\begin{equation}
\small
x_{i}^{\prime}=\operatorname{clip}_{x, \epsilon}\left(x_{i-1}^{\prime}+\frac{\epsilon}{r} \cdot \operatorname{sign}\left(g_{i}\right)\right)
\label{eq:MIM}
\end{equation}

where $g_{i}=\mu g_{i-1}+\frac{\nabla_{x} \ell\left(x_{i-1}^{\prime}, y\right)}{\left\|\nabla_{x} \ell\left( x_{i-1}^{\prime}, y\right)\right\|_{1}}$, and ${\left\|.\right\|_{1}}$ denotes the $L_1$ norm.
\subsection{On techniques to generate soft labels}
Label smoothing is perhaps the most popularly used technique to generate soft labels \cite{2016Label,szegedy2016rethinking,RafaelMuller2019WhenDL,2021Enhancing}. Although it is simple, it has been demonstrated as an effective approach to improve the accuracy of deep learning predictions. For example, Szegedy et al. propose gener-ating soft labels by averaging the one-hot vector and a uniform distribution over labels \cite{szegedy2016rethinking}. Fu et al. study the effect of label smoothing on adversarial training and find that adversarial training with the aid of label smoothing can enhance model robustness against gradient-based attacks \cite{fu2020label}. Wang et al. propose an adaptive label smoothing approach capable of adaptively estimating a target label distribution \cite{wang2021diversifying}. Guo et al. propose a label confusion model (LCM) to improve text classification performance \cite{guo2020label}, which we adapt here for CLDL and generating soft labels for adversarial ensemble prediction.
\section{CLDL Assisted Ensemble Training}
\label{subsec:Our approach}
\begin{figure*}[!htb]
\centering
\includegraphics[width=5.5in]{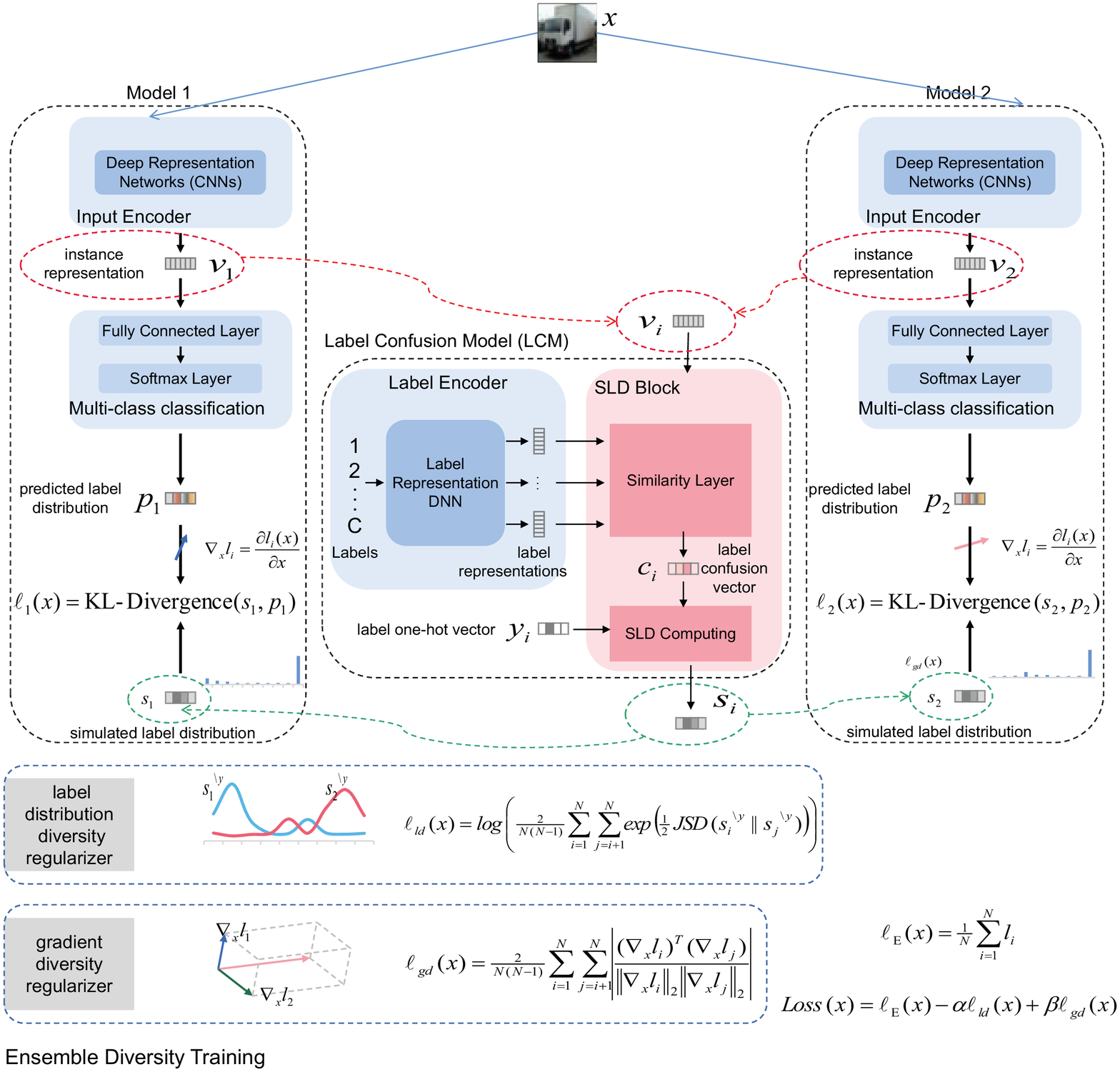}
\caption{The proposed CLDL assisted ensemble training method. Here we take an ensemble model that consists of two member models as an example for ease of illustration. Given an image instance $x$, the label confusion model (LCM) in the middle, which is adapted from \cite{guo2020label}, is used to generate a soft label $s_i$ for the $i$th sub-model. Two types of diversity regularizers that are based on the label distribution given the soft labels and the gradient are combined to generate the finally used ensemble diversity regularizer. See the text in Section \ref{subsec:Our approach} for more details. The LCM module is drawn referring to Fig.1 of \cite{guo2020label}. Note that the model of
\cite{guo2020label} is used for text classification, while here it is adapted to an ensemble model for image classification under adversarial attacks.}
\label{fig:LSD}
\end{figure*}

Here we describe our CLDL based ensemble training approach in detail. We present a pseudo-code implementation of our approach in Algorithm \ref{alg:The LSD training procedure} and a conceptual illustration in Fig.\ref{fig:LSD}. For ease of presentation, we will use an example of an ensemble consisting of $N=2$ member models.
\begin{algorithm}[!htb]
\caption{CLDL assisted ensemble training for an ensemble of $N$ sub-models}
\label{alg:The LSD training procedure}
\begin{algorithmic}[1]
\STATE {Initialization: Initialize $N$ individual models $\{\mathcal{F}_i(x, \theta _i)\}_{i\in [N]}$, where $[N]=\{1,2,\ldots,N\}$, and the label confusion model LCM$(v,y,\theta)$, where $\theta$ denotes the model parameter. Input the training dataset $\mathcal{D}_{train} = \{(x_j, y_j)\}_{j\in [M]}$ to the algorithm. Set the number of training epochs $\mathcal{T}$, the learning rates $\epsilon_i$ and $\epsilon$, $\forall i \in [N]$, the iteration number $\mathcal{IN}$, and the model indicator set $\mathcal{I} = [N]$}.
\FOR{current epoch $t$ = 1 to $\mathcal{T}$}
\FOR{$iter$ = 1 to $\mathcal{IN}$}
\STATE {Sample a mini-batch $\mathcal{D}_{iter}$ from $\mathcal{D}_{train}$}
\FOR{$i$ in $\mathcal{I}$}
\STATE {Obtain the instance representation $\{v_{j}(x_j, \theta _i)\}_{x_j\in \mathcal{D}_{iter}}$ from the outputs of the last Linear layer of $\mathcal{F}_i(x_j, \theta _i)$.}
\STATE {Obtain the predicted label distribution $\{p_{j}(x_j, \theta _i)\}_{x_j\in \mathcal{D}_{iter}}$ from the outputs of soft-max layer of $\mathcal{F}_i(x_j, \theta _i)$.}
\STATE {Obtain the simulated label distribution $\{s_{j}(x_j, \theta _i)\}_{x_j\in \mathcal{D}_{iter}}$ from LCM$(v_j,y_j,\theta)$ by Eqn. (\ref{eq:LCD}).}
\STATE {Calculate loss $\ell_{i} (x_j, \theta _i)_{x_j\in \mathcal{D}_{iter}}$ by Eqn. (\ref{eq:KL-divergence}).}
\STATE {Calculate the gradients of the loss $({\nabla_{x_j} \ell_{i}})_{x_j\in \mathcal{D}_{iter}}$.}
\ENDFOR
\STATE {Calculate the label distribution similarity loss ${\ell_{l d}(x_j)}_{x_j\in \mathcal{D}_{iter}}$ by Eqn. (\ref{eq:label distribution similarity loss})}
\STATE {Calculate the model loss gradient consistency loss ${\ell_{g d}(x_j)}_{x_j\in \mathcal{D}_{iter}}$ by Eqn. (\ref{eq:model loss gradient consistency loss})}
\STATE {Calculate the combined diversity promoting loss ${\ell_{\mathcal{F}, \mathcal{G}}(x_j)}_{x_j\in \mathcal{D}_{iter}}$ by Eqn. (\ref{eq:LSD regularizer})}
\STATE {Obtain the final loss on $\mathcal{D}_{iter}$, denoted by $\mathcal{L}^{iter}_{CLDL}$, using Eqn.(\ref{eq:final learning objective}).}
\FOR{$i$ in $\mathcal{I}$}
\STATE {Update $\theta ^{i}\gets \theta ^{i}- \epsilon_{i}\bigtriangledown_{\theta ^{i}}\mathcal{L}^{iter}_{CLDL}|_{\{\theta _i\}_{i\in[N]}} $}
\ENDFOR
\STATE {Update $\theta \gets \theta - \epsilon \bigtriangledown_{\theta}\mathcal{L}^{iter}_{CLDL}|_{\{\theta\}} $ for LCM$(v,y,\theta)$.}
\ENDFOR
\ENDFOR
\RETURN {The parameters $\theta ^{i}$ and $\theta$, where $i \in [N]$.}
\end{algorithmic}
\end{algorithm}

Our model is mainly composed of two parts: an ensemble of $N$ sub-models ($\{\mathcal{F}_i(x, \theta _i)\}_{i\in [N]}$, where $[N]=\{1,2,\ldots,N\}$) and a label confusion model (LCM) adapted from \cite{guo2020label}. Each sub-model in the ensemble consists of an input convolutional neural network (CNN) encoder followed by a fully connected classifier, which can be any main stream DNN based image classifier. As shown in Fig.\ref{fig:LSD}, an image instance ($x$) is fed into the input-encoder, which generates an image representation $v_{i}$, where $i$ is the sub-model index. Then $v_{i}$ is fed into the fully connected classifier to predict the label distribution of this image. The above operations can be formulated as follows:
\begin{equation}
\small
\begin{aligned}
v_{i} &=\mathcal{F}_{i}^{encoder}(x) \\
p_{i} &= \operatorname{softmax}\left(Wv_{i}+b\right)
\end{aligned}
\label{eq:PLD}
\end{equation}
where $\mathcal{F}_{i}^{encoder}(.)$ is the output of input-encoder of $\mathcal{F}_{i}$ which transforms $x$ to $v_{i}$, $W$ and $b$ are weights and the bias of the fully connect layer that transforms $v_{i}$ to the predicted label distribution (PLD) $p_{i}$.

The LCM consists of two parts: a label encoder and a simulated label distribution (SLD) computation block. The label encoder is a deep neural network used to gener-ate the label representation matrix \cite{guo2020label}. The SLD computation block comprises a similarity layer and an SLD computation layer. The similarity layer takes the label representation matrix and the current instance's representation as inputs, computes the dot product similarities between the image instance and each image class label, then feeds the similarity values into a soft-max activation layer that outputs the label confusion vector (LCV). The LCV captures the conditional dependencies among the image classes through the computed similarities between the instance representation and the label representations. The LCV is instance-dependent, meaning it considers the dependency relationships among all image class labels conditional on a specific image instance. In the following SLD computation layer, the one-hot vector formed hard label $y_i$ is added to the LCV with a controlling parameter $\gamma$, which is then normalized by a soft-max function that generates the SLD. The controlling parameter $\gamma$ decides how much of the one-hot vector will be changed by the LCV. The above operations can be formulated as follows:
\begin{equation}
\small
\begin{aligned}
Vec^{(l)} &=f^{L}(l)=f^{L}\left(\left[l_{1}, l_{2}, \ldots, l_{C}\right]\right) =\left[Vec_{1}^{(l)}, Vec_{2}^{(l)}, \ldots, Vec_{C}^{(l)}\right] \\
c_{i} &=\operatorname{softmax}\left(v_{i}^{\top} Vec^{(l)} W+b\right) \\
s_{i} &=\operatorname{softmax}\left(\gamma y_{i}+c_{i}\right)
\end{aligned}
\label{eq:LCD}
\end{equation}
where $f^L$ is the label encoder function to transfer labels $l=\left[l_{1}, l_{2}, \ldots, l_{C}\right] $ to the label representation matrix $Vec^{(l)}$, $C$ the number of image classes, $f^{L}$ is implemented by an embedding lookup layer followed by a DNN, $c_i$ the LCV and $s_i$ the SLD. The SLD is then viewed as a soft label that replaces the original hard label for model training.

Note that the SLD $s_i$ and the predicted label vector $p_{i}$ are both probability measures. We use the Kullback-Leibler (KL) divergence \cite{1951On} to measure their difference:
\begin{equation}
\small
\begin{aligned}
\ell_i (x) =KL\left(s_{i}, p_{i}\right)=\sum_{c=1}^{C} {s_i}^{c} \log \left(\frac{{s_i}^{c}}{{p_i}^{c}}\right)
\end{aligned}
\label{eq:KL-divergence}
\end{equation}
The LCM is trained by minimizing the above KL divergence, whose value depends on the semantic representation of the image instance $v_i$ and the soft label $s_i$ given by the LCM.
\subsection{Diversity Promoting Loss Design}
\label{subsubsec:LSD Regularization for diversity Measuring}
Here we present our design of the diversity promoting loss used in our approach.
\subsubsection{Soft label diversity}
For an input $(x, y)$ in the training dataset, we define the soft label diversity based on the non-maximal value of the SLD of each sub-model. Specifically, let $s_{i}^{\setminus y}$ be a $(C-1)\times 1$ vector constructed by excluding the maximal value from the SLD corresponding to model ${\mathcal{F}_i(x, \theta)}$. Then we use the Jensen-Shannon divergence (JSD) \cite{ML1997The} to measure the difference between a pair of, say the $i$th and the $j$th member models, in terms of their predicted soft labels, as follows
\begin{equation}
\small
\mbox{JSD}\left(s_{i}^{\setminus y} \| s_{j}^{\setminus y}\right)=\frac{1}{2} \left(KL\left(s_{i}^{\setminus y}, \frac{s_{i}^{\setminus y}+s_{j}^{\setminus y}}{2}\right)+ KL\left(s_{j}^{\setminus y}, \frac{s_{i}^{\setminus y}+s_{j}^{\setminus y}}{2}\right)\right)
\label{eq:JSD}
\end{equation}
From Eqns. (\ref{eq:LCD}) and (\ref{eq:JSD}), we see that JSD$\left(s_{i}^{\setminus y} \| s_{j}^{\setminus y}\right)$ monotonically increases with JSD$\left(c_{i}^{\setminus y} \| c_{j}^{\setminus y}\right)$.
A large JSD indicates a misalignment between the SLDs of the two involved sub-models given the image instance $x$. Given $x$ and an ensemble of $N$ models, we define a loss item as follows
\begin{equation}
\small
\ell_{l d}(x)=\log \left(\frac{2}{N(N-1)} \sum_{i=1}^{N} \sum_{j=i+1}^{N} \exp \left(\mbox{JSD}\left(s_{i}^{\setminus y} \| s_{j}^{\setminus y}\right)\right)\right)
\label{eq:label distribution similarity loss}
\end{equation}
which will be included in the final loss function Eqn.(\ref{eq:final learning objective}) used for training the model ensemble. It plays a vital role in promoting the member models' diversity concerning their pre-dicted soft labels given any input instance.

It is worth noting that we only consider the non-maximal values of the SLDs in Eqn.(\ref{eq:JSD}) following \cite{pang2019improving}. By doing so, promoting the diversity among the sub-models does not affect the en-semble's prediction accuracy for benign inputs. However, it can lower the transferabil-ity of attacks among the sub-models.
\subsubsection{Gradient diversity}
Following \cite{yang2021trs,kariyappa2019improving}, we consider the sub-models' diversity in terms of gradients associated with them. Given an image instance $x$ and an ensemble of $N$ models, we define the gradient diversity loss item as follows
\begin{equation}\label{eq:model loss gradient consistency loss}
\small
\ell_{gd}(x)=\frac{2}{N(N-1)}\sum_{i=1}^{N}\sum_{j=i+1}^{N}CS(\nabla_{x}\ell_{i},\nabla_{x}\ell_{j})=\frac{2}{N(N-1)}\sum_{i=1}^{N}\sum_{j=i+1}^{N}\left|\frac{\left(\nabla_{x}\ell_{i}\right)^{T}\left(\nabla_{x} \ell_{j}\right)}{\left\|\nabla_{x}\ell_{i}\right\|_{2}\left\|\nabla_{x}\ell_{j}\right\|_{2}}\right|
\end{equation}
which will also be included in the final loss function Eqn.(\ref{eq:final learning objective}).
\subsubsection{The Combined Diversity Promoting Loss}
Combining the above soft label and gradient diversity loss items, we propose our CLDL based ensemble diversity loss function. For a pair of member models $(\mathcal{F}$ and $\mathcal{G})$, given an input instance $x$, this diversity promoting loss function is
\begin{equation}
\small
\ell_{\mathcal{F}, \mathcal{G},x} = -\alpha \ell_{l d}(x)+\beta \ell_{g d}(x),
\label{eq:LSD regularizer}
\end{equation}
where $\alpha$, $\beta \ge 0$ are hyper-parameters that balance the effects of the soft label based and the gradient based diversity loss items.
\subsection{CLDL based Ensemble Model Training}
\label{subsubsec:LSD Diversity Training Strategies for Ensemble Models}
We train our ensemble model by minimizing the training loss function defined as follows
\begin{equation}
\small
\operatorname{Loss}(x)=\ell_{\mathrm{E}}(x) + \ell_{\mathcal{F}, \mathcal{G},x} = (\frac{1}{N} \sum_{i=1}^{N} \ell_{i}) -\alpha \ell_{l d}(x)+\beta \ell_{g d}(x)
\label{eq:final learning objective}
\end{equation}
where $ \ell_{\mathrm{E}}(x)=\frac{1}{N} \sum_{i=1}^{N} \ell_{i}$ refers to the average of the KL-divergence losses of the member models. See Fig.\ref{fig:LSD} for the definition of the KL loss of the member models.
By minimizing the above loss function, we simultaneously learn the soft labels given by each sub-model, promote the diversity among the sub-models in terms their predicted soft labels and their gradients, and minimize the KL-divergence loss of each sub-model.
\section{Experiments}
\label{sec:Simulation}
\subsection{Datasets and Competitor Methods}
We conducted our experiments on the widely-used image datasets MNIST \cite{deng2012mnist}, Fasion-MNIST (F-MNIST) \cite{xiao2017fashion}, and CIFAR-10 \cite{krizhevsky2009learning}. For each dataset, we used its training set for ensemble training. We set the hyper-parameters of our algorithm based on 1,000 test images randomly sampled from the testing set and used the remaining data in the testing set for performance evaluation.

We compared the performance of our algorithm with competitor methods, including a baseline method that trains the model ensemble without the use of any defense mechanism and four popularly used ensemble training methods: the adaptive diversity promoting (ADP) algorithm, the gradient alignment loss (GAL) method, the diversi-fying vulnerabilities for enhanced robust generation of ensembles (DVERGE) meth-od, and the transferability reduced smooth (TRS) method. We used ResNet-20 as the basic model structure of the sub-models and averaged the output probabilities given by the softmax layer of the member models to yield the final prediction.
\subsection{Optimizer for Training}
\label{subsub:Training Setup}
We use Adam \cite{kingma2014adam} as the optimizer for ensemble training with an initial learning rate of $10^{-3}$, and a weight decaying parameter of $10^{-4}$. For our CLDL-based approach, we trained the ensemble for 200 epochs, multiplied the learning rate by 0.1 twice at the 100th and 150th epochs, respectively. We set the batch size to 128 and used normalization, random cropping, and flipping-based data augmentations for dataset CIFAR-10. We considered two ensemble sizes, 3 and 5, in our experiments. To make a fair comparison, we trained ADP, GAL, DVERGE, and TRS under a similar training setup described above. We used the AdverTorch \cite{ding2019advertorch} library for simulating adversarial attacks.
\subsection{White-box Attacks}
We considered four basic white-box adversarial attacks, namely FGSM, BIM, MIM, and PGD for simulating black-box attacks used in our experiments. For each attack type, we considered four different perturbation scales ($\epsilon$) ranging from 0.01 to 0.04. We set the number of attack iterations at 10 and set the step size to be $\epsilon$/5 for BIM, MIM and PGD. Each experiment was run five times independently, and the results were averaged for performance comparison. We simulated the white-box attacks by treating the whole ensemble, other than one of the individual sub-models, as the target model to be attacked.
\subsection{Black-box Attacks}
We considered black-box attacks, in which the attacker has no knowledge about the target model, including its architecture and parameters. The attacker designs adversarial examples based on several surrogate models. We simulated black-box attacks with our ensemble model as the target by creating white-box adversarial attacks based on a surrogate ensemble model that has the same architecture as the true target ensemble and is trained on the same dataset using the same training routine.

We trained the surrogate ensemble model consisting of 3- or 5-member sub-models by minimizing a standard ensemble cross-entropy loss function. For each type of attack mentioned above, we evaluated the robustness of the involved training meth-ods under black-box attacks with four different perturbation scales ($\epsilon$) 0.01,0.02,0.03 and 0.04. We set the number of attack iterations at 10, and the step size at $\epsilon$/5 for BIM, MIM and PGD based attacks. Following \cite{yang2020dverge}, we generate adversarial examples using the cross-entropy loss and the CW loss \cite{carlini2017towards}.
\subsection{Experimental Results for Black-box Adversarial Attacks}
\label{subsec:Experimental Results}
In our experiments, we used classification accuracy as the performance metric, which is the ratio of the number of correctly predicted adversarial instances to the total number of adversarial instances. We conducted random re-trainings of the model in our experiments, and the reported values are averages of multiple $(>3)$ independent tests. Our code is open-sourced to support reproducibility of these results.
\subsubsection{CIFAR-10}
Here we present our experimental results on CIFAR-10 in Tables \ref{tab:BlackboxAdversarialAttacksCIFAR_3} and \ref{tab:BlackboxAdversarialAttacksCIFAR_5}. Note that, in all tables shown below, $CLDL_{a,b,c}$ denotes our CLDL based algorithm with hyper-parameters, namely $\gamma$ in Eqn.(\ref{eq:LCD}), $\alpha$ and $\beta$ in Eqn.(\ref{eq:final learning objective}), set to be $a$, $b$, and $c$, respectively. $\epsilon$ refers to the perturbation scale of the attack.
As is shown, our CLDL based algorithm performs best for almost all attacks considered, compared to the other methods, especially when the perturbation scale is large.
\begin{table}[!htb]
\caption{\label{tab:BlackboxAdversarialAttacksCIFAR_3} Classification accuracy ($\%$) on the CIFAR-10 dataset for four types of black-box attacks. The ensemble consists of 3 ReNets-20 member models. $\epsilon$ refers to the perturbation scale for the attack.}
\centering
\setlength{\tabcolsep}{0.6mm}{
\small
\begin{tabular}{@{}c|c|c|c|c|c|c|c@{}}
\toprule
\multicolumn{1}{l|}{\textbf{CIFAR-10}}     & $\epsilon$     & ADP     & GAL     & DVERGE  & TRS     & $CLDL_{4,2,4}$       & $CLDL_{4,0.5,4}$  \\  \hline
\multirow{4}{*}{BIM}            & 0.01 & 45.43          & 92.61          & 89.66          & 87.86 & \textbf{92.77}  &\textbf{92.77}           \\
                                & 0.02 & 9.74           & 83.59          & 75.41          & 71.17 & \textbf{85.21}  &83.66                    \\
                                & 0.03 & 2.01           & 73.02          & 59.24          & 53.74 & \textbf{75.76}  &72.97                    \\
                                & 0.04 & 0.47           & 62.32          & 42.27          & 37.87 & \textbf{66.47}  &62.38                    \\ \hline
\multirow{4}{*}{FGSM}           & 0.01 & 67.03          & 93.47          & 91.20          & 90.37 & \textbf{93.49}  &93.47                    \\
                                & 0.02 & 40.58          & 84.62          & 78.68          & 77.59 & \textbf{86.21}  &85.48                    \\
                                & 0.03 & 26.41          & 75.93          & 65.37          & 64.12 & \textbf{77.94}  &76.04                    \\
                                & 0.04 & 18.25          & 66.15          & 51.40          & 51.04 & \textbf{69.93}  &66.95                    \\ \hline
\multirow{4}{*}{MIM}            & 0.01 & 41.65          & 91.11          & 87.52          & 85.39 & \textbf{91.29}  &91.08                    \\
                                & 0.02 & 7.37           & 76.92          & 65.85          & 61.80 & \textbf{78.98}  &77.01                    \\
                                & 0.03 & 1.13           & 59.75          & 40.35          & 38.01 & \textbf{63.84}  &59.89                    \\
                                & 0.04 & 0.30           & 43.48          & 19.77          & 20.24 & \textbf{48.27}  &43.06                    \\ \hline
\multirow{4}{*}{PGD}            & 0.01 & 46.20          & \textbf{92.41} & 89.36          & 88.06 & 92.18           &92.25                    \\
                                & 0.02 & 9.23           & 83.46          & 76.08          & 72.07 & \textbf{84.81}  &83.68                    \\
                                & 0.03 & 1.55           & 74.67          & 61.28          & 55.12 & \textbf{77.34}  &75.04                    \\
                                & 0.04 & 0.35           & 65.46          & 44.72          & 39.23 & \textbf{70.64}  &67.09                    \\ \bottomrule
\end{tabular}}
\end{table}
\begin{table}[!htb]
\caption{\label{tab:BlackboxAdversarialAttacksCIFAR_5} Classification accuracy ($\%$) on the CIFAR-10 dataset for four types of black-box attacks. The ensemble consists of 5 ReNets-20 member models.}
\centering
\setlength{\tabcolsep}{0.6mm}{
\small
\begin{tabular}{@{}c|c|c|c|c|c|c|c@{}}
\toprule
\multicolumn{1}{l|}{\textbf{CIFAR-10}}     & $\epsilon$     & ADP     & GAL     & DVERGE  & TRS     & $CLDL_{4,2,4}$       & $CLDL_{4,0.5,4}$\\  \hline
\multirow{4}{*}{BIM}            & 0.01 & 45.28          & 92.22          & 93.32          & 91.17 & \textbf{93.73}  & 92.89                \\
                                & 0.02 & 9.50           & 81.69          & 85.27          & 80.15 & \textbf{85.71}  & 84.58                \\
                                & 0.03 & 1.79           & 69.68          & 75.72          & 67.93 & \textbf{75.93}  & 75.56                \\
                                & 0.04 & 0.44           & 57.21          & 64.59          & 55.95 & {65.39}  & \textbf{65.8}         \\ \hline
\multirow{4}{*}{FGSM}           & 0.01 & 68.49          & 93.33          & 94.22          & 92.65 & \textbf{94.47}  & 93.54                \\
                                & 0.02 & 43.02          & 84.13          & 86.79          & 83.30 & \textbf{87.06}  & 85.92                \\
                                & 0.03 & 27.76          & 74.15          & 77.56          & 72.24 & \textbf{78.79}  & 78.3                 \\
                                & 0.04 & 19.36          & 64.32          & 67.10          & 61.08 & \textbf{69.93}  & 69.68                \\ \hline
\multirow{4}{*}{MIM}            & 0.01 & 41.77          & 90.50          & 91.92          & 89.53 & \textbf{92.17}  & 91.4                 \\
                                & 0.02 & 7.42           & 73.95          & \textbf{79.69} & 73.33 & 79.33           & 78.67                \\
                                & 0.03 & 1.16           & 55.09          & 62.87          & 54.63 & \textbf{63.24}  & 62.93                \\
                                & 0.04 & 0.25           & 37.87          & 44.19          & 38.76 & \textbf{47.16}  & 46.62                \\ \hline
\multirow{4}{*}{PGD}            & 0.01 & 46.50          & 92.22          & \textbf{93.34} & 91.29 & 92.92           & 92.02                \\
                                & 0.02 & 9.43           & 81.85          & 85.60          & 80.24 & \textbf{85.92}  & 84.45                \\
                                & 0.03 & 1.54           & 71.89          & 77.59          & 68.25 & \textbf{78.07}  & 76.87                \\
                                & 0.04 & 0.29           & 61.60          & 68.53          & 55.45 & \textbf{70.05}  & 69.78                \\ \bottomrule
\end{tabular}}
\end{table}

We also investigate the effects of the soft label diversity based loss and the gradient diversity based one on the performance of our algorithm. See the result in Table \ref{tab:BlackboxAdversarialAttacksCIFAR_3_hyper}. As is shown, $CLDL_{4,2,4}$ gives the best results. By comparing the result of $CLDL_{4,2,4}$ to that of $CLDL_{4,0.5,4}$, we find a performance gain given by the soft label diversity based loss. By comparing the result of $CLDL_{4,2,4}$ to that of $CLDL_{4,2,0}$, we verify the contribution of the gradient diversity based loss.
\begin{table*}[!htb]
\centering
\caption{\label{tab:BlackboxAdversarialAttacksCIFAR_3_hyper} Classification accuracy $(\%)$ given by an ensemble model consisting of 3 ReNets-20 member models trained with our CLDL based algorithm with different hyper-parameter settings against black-box attacks on the CIFAR-10 dataset.}
\setlength{\tabcolsep}{0.6mm}{
\tiny
\begin{tabular}{c|c|c|c|c|c|c|c|c|c}
\hline
\textbf{CIFAR-10}          & $\epsilon$  & $CLDL_{4,0,0}$  & $CLDL_{4,1,0}$ & $CLDL_{4,2,0}$ & $CLDL_{4,4,0}$ &  $CLDL_{4,1,2}$  & $CLDL_{4,2,2}$ & $CLDL_{4,0.5,4}$ & $CLDL_{4,2,4}$  \\ \hline
\multirow{4}{*}{BIM}  & 0.01 & 89.97                   & 89.66                   & 89.82                   & 89.78                   & 92.76                   & 92.62                   & \textbf{92.77}          & \textbf{92.77}          \\
                      & 0.02 & 74.01                   & 74.18                   & 74.2                    & 74.42                   & 83.43                   & 82.97                   & 83.66                   & \textbf{85.21}          \\
                      & 0.03 & 57.1                    & 56.62                   & 56.78                   & 56.92                   & 73.02                   & 72.64                   & 72.97                   & \textbf{75.76}          \\
                      & 0.04 & 40.11                   & 39.63                   & 40.63                   & 40.53                   & 61.8                    & 61.69                   & 62.38                   & \textbf{66.47}          \\ \hline
\multirow{4}{*}{FGSM} & 0.01 & 91.53                   & 91.07                   & 91.01                   & 91.12                   & \textbf{93.5}           & 93.2                    & 93.47                   & 93.49                   \\
                      & 0.02 & 78.64                   & 78.63                   & 78.78                   & 78.51                   & 84.79                   & 84.62                   & 85.48                   & \textbf{86.21}          \\
                      & 0.03 & 64.83                   & 64.37                   & 65.22                   & 64.37                   & 76.05                   & 76.24                   & 76.04                   & \textbf{77.94}          \\
                      & 0.04 & 52.36                   & 51.87                   & 53.11                   & 52.16                   & 66.55                   & 67.42                   & 66.95                   & \textbf{69.93}          \\ \hline
\multirow{4}{*}{MIM}  & 0.01 & 87.64                   & 87.18                   & 87.39                   & 87.26                   & 91.22                   & 90.73                   & 91.08                   & \textbf{91.29}          \\
                      & 0.02 & 63.63                   & 63.24                   & 63.79                   & 63.86                   & 76.16                   & 76.3                    & 77.01                   & \textbf{78.98}          \\
                      & 0.03 & 39.03                   & 38.62                   & 39.29                   & 38.82                   & 58.9                    & 59.25                   & 59.89                   & \textbf{63.84}          \\
                      & 0.04 & 20.84                   & 20.29                   & 21.5                    & 21.42                   & 41.79                   & 43.07                   & 43.06                   & \textbf{48.27}          \\ \hline
\multirow{4}{*}{PGD}  & 0.01 & 90.24                   & 89.74                   & 89.76                   & 89.75                   & \textbf{92.26}          & 92.09                   & 92.25                   & 92.18                   \\
                      & 0.02 & 75.54                   & 75.66                   & 75.48                   & 75.69                   & 83.25                   & 82.81                   & 83.68                   & \textbf{84.81}          \\
                      & 0.03 & 60.63                   & 60.03                   & 60.55                   & 60.13                   & 74.53                   & 74.45                   & 75.04                   & \textbf{77.34}          \\
                      & 0.04 & 46.97                   & 46.2                    & 46.68                   & 46.41                   & 66.03                   & 65.95                   & 67.09                   & \textbf{70.64}          \\ \hline
\end{tabular}}
\end{table*}
\begin{table}[!htb]
\caption{\label{tab:BlackboxAdversarialAttacksMNIST_3} Classification accuracy ($\%$) given by an ensemble model consisting of 3 LeNet-5 member models trained with our CLDL based algorithm against black-box attacks on the MNIST dataset. }
\centering
\setlength{\tabcolsep}{0.6mm}{
\tiny
\begin{tabular}{@{}c|c|c|c|c|c|c|c@{}}
\toprule
\multicolumn{1}{l|}{\textbf{MNIST}}     & $\epsilon$     & ADP     & GAL     & DVERGE  & TRS     & $CLDL_{3,4,4}$  & $CLDL_{3,2,1}$    \\  \hline
\multirow{4}{*}{BIM}  & 0.1  & 90.18 & 87.34 & 90.24  & 92.5          & \textbf{94.48} & 94.22          \\
                      & 0.15 & 60.38 & 55.61 & 61.21  & 76.63         & \textbf{85.64} & 81.42          \\
                      & 0.2  & 23.23 & 28.5  & 21.17  & 46.16         & \textbf{65.11} & 51.59          \\
                      & 0.25 & 5.32  & 11.06 & 2.53   & 17.42         & \textbf{32.81} & 22.95          \\ \hline
\multirow{4}{*}{FGSM} & 0.1  & 93.29 & 90.77 & 93.51  & 94.55             & 95.43          & \textbf{95.5}  \\
                      & 0.15 & 79.58 & 69.94 & 80.75  & 86.21         & \textbf{89.82} & 88.56          \\
                      & 0.2  & 52.99 & 47.35 & 55.98  & 70.64         & \textbf{78.39} & 72.73          \\
                      & 0.25 & 27.38 & 30.13 & 27.78  & {48.1}               & \textbf{57.51} & 46.77          \\ \hline
\multirow{4}{*}{MIM}  & 0.1  & 90.21 & 85.82 & 90.52  & 92.31         & \textbf{94.25} & 94.07          \\
                      & 0.15 & 63.05 & 53.72 & 63.67  & 76.81         & \textbf{85.31} & 81.56          \\
                      & 0.2  & 24.69 & 27.49 & 23.88  & 46.83         & \textbf{65.1}  & 51.33          \\
                      & 0.25 & 5.58  & 10.64 & 3.16   & 16.58         & \textbf{30.44} & 21.34          \\ \hline
\multirow{4}{*}{PGD}  & 0.1  & 89.66 & 84.87 & 89.82  & 91.91             & 93.75          & \textbf{93.84} \\
                      & 0.15 & 56.83 & 47.69 & 57.86  & 73.01         & \textbf{83.5}  & 78.92          \\
                      & 0.2  & 19.42 & 21.69 & 17.32  & 39.19         & \textbf{58.94} & 46.91          \\
                      & 0.25 & 3.06  & 6.44  & 1.05   & 11.71         & \textbf{24.81} & 16.65          \\ \bottomrule
\end{tabular}}
\end{table}
\begin{table}[!htb]
\caption{\label{tab:BlackboxAdversarialAttacksMNIST_5} Robust accuracy ($\%$) of an ensemble of 5 LeNet-5 models against black-box attacks on the MNIST dataset}
\centering
\setlength{\tabcolsep}{0.6mm}{
\tiny
\begin{tabular}{@{}c|c|c|c|c|c|c|c@{}}
\toprule
\multicolumn{1}{l|}{\textbf{MNIST}}         & $\epsilon$  & ADP     & GAL     & DVERGE  & TRS             & $CLDL_{3,4,4}$ & $CLDL_{3,2,1}$   \\  \hline
\multirow{4}{*}{BIM}                & 0.1    & 88.43     & 90.26     & 89.49     & {94.01}    & \textbf{95.01}                  & 93.98 \\
                                    & 0.15   & 53.68     & 66.19     & 60.83     & {82.77}    & \textbf{87.18}          & 78.76 \\
                                    & 0.2    & 18.98     & 34.1      & 23.04     & {55.12}    & \textbf{68.76}          & 51.46 \\
                                    & 0.25   & 2.21      & 12.02     & 4.01      & 22.2              & \textbf{40.58}          & 25.3  \\ \hline
\multirow{4}{*}{FGSM}               & 0.1    & 92.23     & 93.03     & 92.88     & 95.41             & 95.9                    & 95.62 \\
                                    & 0.15   & 75.07     & 79.74     & 80.23     & 88.85    & \textbf{89.86}                   & 88.43 \\
                                    & 0.2    & 46.26     & 58.21     & 54.71     & 76.11    & \textbf{76.95}                   & 71.07 \\
                                    & 0.25   & 22.41     & 35.67     & 29.5      & \textbf{54.6}     & 52.6                    & 44.93 \\ \hline
\multirow{4}{*}{MIM}                & 0.1    & 89.06     & 89.62     & 89.84     & 93.96    & \textbf{94.68}                   & 93.96 \\
                                    & 0.15   & 56.8      & 66.79     & 63.58     & 82.61    & \textbf{85.96}                   & 78.68 \\
                                    & 0.2    & 21.1      & 35.18     & 26.19     & 56.63    & \textbf{65.32}          & 50.11 \\
                                    & 0.25   & 3.02      & 12.37     & 5.33      & 22.55    & \textbf{34.31}          & 22.36 \\ \hline
\multirow{4}{*}{PGD}                & 0.1    & 87.83     & 88.69     & 89.11     & 93.53    & \textbf{94.4}                    & 93.49 \\
                                    & 0.15   & 50.51     & 60.26     & 57.6      & {81.19}    & \textbf{85.08}          & 74.77 \\
                                    & 0.2    & 15.33     & 27.47     & 19.45     & {50.15}    & \textbf{63.35}          & 45.58 \\
                                    & 0.25   & 0.97      & 6.64      & 2.19      & 17.06             & \textbf{31.88}          & 18.38 \\ \bottomrule
\end{tabular}}
\end{table}
\begin{table}[!htb]
\caption{\label{tab:BlackboxAdversarialAttacksF-MNIST_3} Classification accuracy ($\%$) of an ensemble of 3 LeNet-5 models against black-box attacks on the F-MNIST dataset.}
\centering
\setlength{\tabcolsep}{0.6mm}{
\small
\begin{tabular}{@{}c|c|c|c|c|c|c@{}}
\toprule
\multicolumn{1}{l|}{\textbf{F-MNIST}}     & $\epsilon$     & ADP     & GAL     & DVERGE  & TRS & $CLDL_{3,2,4}$   \\  \hline
\multirow{4}{*}{BIM}  & 0.08 & 38.38 & 54.43          & 39.42  & \textbf{54.71} & 54.19                   \\
                      & 0.1  & 28.39 & 44.37          & 28.16  & 44.61          & \textbf{44.99}          \\
                      & 0.15 & 10.55 & 22.63          & 10.58  & 25.06          & \textbf{27.84}          \\
                      & 0.2  & 1.78  & 7.10           & 2.69   & 10.95          & \textbf{13.98}          \\ \hline
\multirow{4}{*}{FGSM} & 0.08 & 48.25 & 62.77          & 52.39  & \textbf{62.83} & 61.99                          \\
                      & 0.1  & 40.09 & \textbf{54.22} & 42.80  & 53.64          & 53.87                   \\
                      & 0.15 & 23.24 & 36.09          & 25.41  & 37.63          & \textbf{38.49}          \\
                      & 0.2  & 9.82  & 15.85          & 13.33  & 23.45          & \textbf{25.67}          \\ \hline
\multirow{4}{*}{MIM}  & 0.08 & 38.24 & 52.83          & 39.58  & \textbf{53.46} & 52.88                   \\
                      & 0.1  & 28.41 & 42.44          & 28.19  & 43.10          & \textbf{43.41}          \\
                      & 0.15 & 9.17  & 19.61          & 9.98   & 23.01          & \textbf{25.20}          \\
                      & 0.2  & 1.01  & 3.44           & 2.33   & 7.84           & \textbf{9.75}           \\ \hline
\multirow{4}{*}{PGD}  & 0.08 & 37.74 & 52.34          & 39.17  & \textbf{52.86} & 51.65                   \\
                      & 0.1  & 27.75 & 41.71          & 28.15  & \textbf{42.38} & 42.23                   \\
                      & 0.15 & 8.69  & 18.78          & 9.67   & 22.44          & \textbf{24.75}          \\
                      & 0.2  & 1.08  & 3.52           & 2.24   & 8.25           & \textbf{10.91}          \\ \bottomrule
\end{tabular}}
\end{table}
\begin{table}[!htb]
\caption{\label{tab:BlackboxAdversarialAttacksF-MNIST_5} Classification accuracy ($\%$) of an ensemble of 5 LeNet-5 models on the F-MNIST dataset}
\centering
\setlength{\tabcolsep}{0.6mm}{
\small
\begin{tabular}{@{}c|c|c|c|c|c|c@{}}
\toprule
\multicolumn{1}{l|}{\textbf{F-MNIST}}         & $\epsilon$  & ADP     & GAL     & DVERGE  & TRS             & $CLDL_{3,2,4}$ \\  \hline
\multirow{4}{*}{BIM}                & 0.08 & 38.01 & 52.62          & 41.36  & 60.30          & \textbf{61.37}         \\
                                    & 0.1  & 28.57 & 42.47          & 29.67  & 51.10          & \textbf{51.72}         \\
                                    & 0.15 & 10.83 & 23.04          & 11.15  & 33.67          & \textbf{35.46}         \\
                                    & 0.2  & 2.01  & 8.86           & 2.63   & 20.48          & \textbf{24.58}         \\ \hline
\multirow{4}{*}{FGSM}               & 0.08 & 48.36 & 62.53          & 54.13  & 65.65          & \textbf{67.90}         \\
                                    & 0.1  & 39.94 & 54.14          & 44.80  & 58.09          & \textbf{59.78}         \\
                                    & 0.15 & 23.00 & 36.65          & 27.28  & 43.66          & \textbf{45.69}         \\
                                    & 0.2  & 9.98  & 19.49          & 14.44  & 31.91          & \textbf{32.71}         \\ \hline
\multirow{4}{*}{MIM}                & 0.08 & 38.22 & 52.16          & 41.69  & 58.93          & \textbf{59.18}         \\
                                    & 0.1  & 28.65 & 42.23          & 30.20  & 49.41          & \textbf{49.48}         \\
                                    & 0.15 & 9.75  & 20.97          & 11.27  & 31.79          & \textbf{33.15}         \\
                                    & 0.2  & 1.26  & 4.94           & 2.12   & 17.11          & \textbf{18.71}         \\ \hline
\multirow{4}{*}{PGD}                & 0.08 & 37.76 & 51.56          & 41.11  & \textbf{59.06} & 58.28                  \\
                                    & 0.1  & 27.86 & 41.03          & 29.50  & \textbf{49.18} & 47.99                  \\
                                    & 0.15 & 9.39  & 20.45          & 10.14  & 31.27          & \textbf{32.97}         \\
                                    & 0.2  & 1.39  & 5.99           & 1.97   & 18.00          & \textbf{20.38}         \\ \bottomrule
\end{tabular}}
\end{table}
\subsubsection{MNIST}
In Tables \ref{tab:BlackboxAdversarialAttacksMNIST_3} and \ref{tab:BlackboxAdversarialAttacksMNIST_5}, we show the classification accuracy  ($\%$) results for an ensemble of 3 and 5 LeNet-5 member models \cite{lecun1989backpropagation} on the MNIST dataset. We find that again our algorithm outperforms its competitors significantly.
\subsubsection{F-MNIST}In Tables \ref{tab:BlackboxAdversarialAttacksF-MNIST_3} and \ref{tab:BlackboxAdversarialAttacksF-MNIST_5}, we present results associated with the F-MNIST dataset. As is shown, among all methods involved, our CLDL algorithm ranks number 1 for 10 times. TRS and GAL have 5 times and one time to rank number 1, respectively.
\section{Conclusion}
\label{sec:CONCLUSION}
In this paper, we proposed a novel adversarial ensemble training approach that leverages conditional label dependency learning. In contrast to existing methods that encode image classes with one-hot vectors, our algorithm can learn and exploit the conditional relationships between labels during member model training. Experimental results demonstrate that our approach is more robust against black-box adversarial attacks than state-of-the-art methods.
\section*{Acknowledgment}
This work was supported by Research Initiation Project (No.2021KB0PI01) and Ex-ploratory Research Project (No.2022RC0AN02) of Zhejiang Lab.

\bibliographystyle{IEEEtran}
\bibliography{mybibfile}
\end{document}